\documentclass[11pt,letterpaper]{article}

\usepackage[utf8]{inputenc}
\usepackage[export]{adjustbox}
\usepackage{float}
\usepackage[english]{babel}
\usepackage{algorithm}
\usepackage[noend]{algpseudocode}
\usepackage{comment}

\usepackage{hyperref}
\usepackage{graphicx}
\usepackage{amsmath,amssymb,mathrsfs,amsthm}
\usepackage{wrapfig}
\usepackage{bm,bbding}
\usepackage{xspace}
\usepackage{xcolor}
\usepackage{float}
\usepackage{subfig}
\usepackage{multirow}

\usepackage{tikz}
\usepackage{pgfplots}
\usepgfplotslibrary{colorbrewer}
\usepgfplotslibrary{groupplots} 
\usetikzlibrary{pgfplots.groupplots}
\usetikzlibrary{matrix,arrows,decorations.pathmorphing}
\usepackage{pgfplots,pgfplotstable}
\usepgfplotslibrary{fillbetween}

    \pgfplotsset{
        colormap/Set1-5,
        colormap/Set2-5,
    }

\usepackage{textpos}
\usepackage{pgfplotstable}
\usepackage{array}
\usepackage{authblk}
\usepackage{multirow}
\usepackage{booktabs}
\usepackage{diagbox}

\usepackage{rays_defs_18}

\long\def\comment#1{}

\usepackage{hyperref}
\hypersetup{
    bookmarks=true,         
    unicode=false,          
    pdftoolbar=true,        
    pdfmenubar=true,        
    pdffitwindow=false,     
    pdfstartview={FitH},    
    pdfauthor={Reinhard Heckel},     
    pdfsubject={Subject},   
    pdfcreator={Reinhard Heckel},   
    pdfproducer={Producer}, 
    pdfnewwindow=true,      
    colorlinks=true,       
    linkcolor=red,          
    citecolor=green,        
    filecolor=magenta,      
    urlcolor=cyan           
}

\usepackage[
backend=biber,
url=false,isbn=false,doi=false,
date=year,
hyperref=auto,
style=alphabetic,
natbib=true,
sorting=nty,
firstinits=true,
maxnames=2, maxbibnames=10,
]{biblatex}
\bibliography{./bibliography.bib} 

\usepackage{bm}
\newcommand\vtheta{{\bm \theta}}

\begin{document}

\begin{center}

{\bf{\LARGE{
Image-to-Image MLP-mixer for Image Reconstruction
}}}

\vspace*{.2in}

{\large{
\begin{tabular}{cccc}
Youssef Mansour$^\ast$, Kang Lin$^\ast$, and Reinhard Heckel$^{\ast,\dagger}$
\end{tabular}
}}

\vspace*{.05in}

\begin{tabular}{c}
$^\ast$Dept. of Electrical and Computer Engineering, Technical University of Munich \\
$^\dagger$Dept. of Electrical and Computer Engineering, Rice University \\
\end{tabular}

\vspace*{.1in}


\vspace*{.1in}

\end{center}

\begin{abstract}
Neural networks are highly effective tools for image reconstruction problems such as denoising and compressive sensing. 
To date, neural networks for image reconstruction are almost exclusively convolutional. The most popular architecture is the U-Net, a convolutional network with a multi-resolution architecture. 
In this work, we show that a simple network based on the multi-layer perceptron (MLP)-mixer enables state-of-the art image reconstruction performance without convolutions and without a multi-resolution architecture, provided that the training set and the size of the network are moderately large. 
Similar to the original MLP-mixer, the image-to-image MLP-mixer is based exclusively on MLPs operating on linearly-transformed image patches. Contrary to the original MLP-mixer, we incorporate structure by retaining the relative positions of the image patches. 
This imposes an inductive bias towards natural images which enables the image-to-image MLP-mixer to learn to denoise images based on fewer examples than the original MLP-mixer. 
Moreover, the image-to-image MLP-mixer requires fewer parameters to achieve the same denoising performance than the U-Net and its parameters scale linearly in the image resolution instead of quadratically as for the original MLP-mixer. If trained on a moderate amount of examples for denoising, the image-to-image MLP-mixer outperforms the U-Net by a slight margin. It also outperforms the vision transformer tailored for image reconstruction and classical un-trained methods such as BM3D, making it a very effective tool for image reconstruction problems.
\end{abstract}

\section{Introduction}
Deep neural networks have emerged as highly successful tools for image and signal reconstruction, restoration, and manipulation.  
They achieve state-of-the-art image quality on tasks like denoising, super-resolution, image reconstruction from few and noisy measurements, and image generation. 

Current state-of-the-art image reconstruction networks are convolutional. 
Convolutional neural networks (CNNs) achieve better denoising image quality than classical methods such as BM3D 
\citep{zhang_GaussianDenoiserResidual_2017,brooks_UnprocessingImagesLearned_2019}. 
They also perform excellent on many other imaging problems including computed tomography~\citep{mccann_ConvolutionalNeuralNetworks_2017} and accelerated magnetic resonance imaging (MRI)~\citep{zbontar_FastMRIOpenDataset_2018}. For example, all top-performing methods at the FastMRI competition, a challenge for accelerated magnetic resonance imaging~\citep{zbontar_FastMRIOpenDataset_2018,knoll_AdvancingMachineLearning_2020}, are CNNs. 


For the related problem of \emph{image classification}, CNNs are also state-of-the-art. However, recent work has shown that new non-convolutional networks can perform comparable when trained on huge datasets. 
For instance, the vision transformer~\citep{dosovitskiy2021an} is an attention-based architecture without convolutions that achieves excellent classification accuracy when pre-trained on very large datasets. 
Most recently, networks solely based on multi-layer perceptrons (MLPs) were proposed, including the MLP-mixer~\citep{tolstikhin_MLPMixerAllMLPArchitecture_2021,liu_PayAttentionMLPs_2021a,chen_CycleMLPMLPlikeArchitecture_2021}. 
Trained on a huge dataset, the MLP-mixer performs almost as well as the best convolutional architectures while having lower computational costs at inference.

Non-convolutional architectures such as the ViT and MLP-mixer impose a lower inductive bias than CNNs. This inductive bias enables CNNs to perform well when little to moderate amounts of training data are available, but might limit performance if abundant data is available. The low inductive bias of the ViT and MLP-mixer causes them to perform very poorly when little data is available.


Motivated by this development, and by the simplicity of the MLP-mixer, we propose and study a variant of the MLP-mixer for image reconstruction tasks, that has moderate inductive bias, with the premise that such a network can perform well when trained on few data, and also gives better image quality than convolutional networks if trained on sufficiently large data sets. 

The architecture of the image-to-image MLP-mixer is depicted in Figure~\ref{fig:mixer}. 
The image-to-image MLP-mixer differs from the original MLP mixer in that it retains the relative positions of the patches, which leads to significantly better performance for image reconstruction tasks. 

Our results show that the image-to-image mixer can outperform a state-of-the-art image reconstruction architecture, the U-Net~\cite{ronneberger_UNetConvolutionalNetworks_2015}, by a small margin. 
We show that the gap in performance between the image-to-image mixer and a U-Net increases with the number of training images and the model size (see Figures~\ref{fig:example} and ~\ref{fig:cs_accuracy}).  We also show that, even in the regime of relatively few training images, the image-to-image MLP-mixer slightly outperforms a U-Net of similar size in image quality, both for denoising images perturbed with Gaussian noise, denoising images perturbed by real-world camera noise, and for compressed sensing reconstruction in magnetic resonance imaging. 
Phrased differently, to achieve the same denoising performance, the image-to-image MLP-mixer requires fewer parameters (see Figure~\ref{fig:example}). Our image-to-image MLP-mixer significantly outperforms the original MLP-mixer and a vision transformer tailored to image-to-image tasks, and BM3D, a classical un-trained denoising algorithm at denoising. 


\section{Image-to-image MLP-mixer network architecture}
%

\begin{figure}[t]
\begin{center}
\includegraphics[width=1\textwidth]{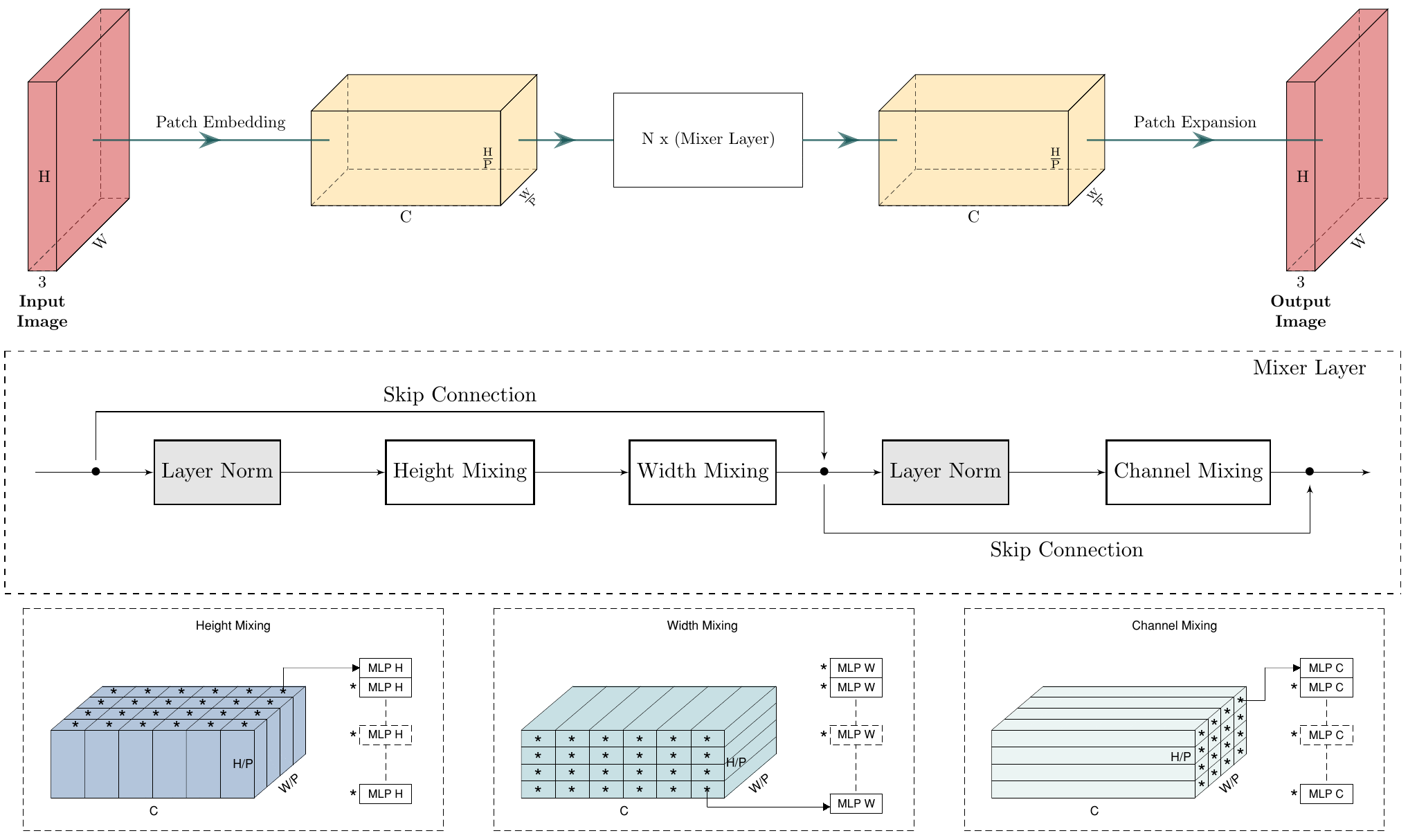}
\end{center}
%
\caption{
\label{fig:mixer}
An illustration of the image-to-image MLP-mixer: First, the image is divided into patches which are transformed with a linear layer to a volume of size $W/P \times H/P \times C$. This volume is transformed with 3D mixer-blocks (16 in our standard architecture), and finally projected back with a linear transformation to an image. 
There are many potential choices for the mixer block, our default one mixes across channels, width, and height separately. 
For example, mixing in channel dimension means applying the same MLP to each of the $H/P \cdot W/P$-many vectors in channel direction. The mixer block we propose retains the relative order of the patches. 
}
\end{figure}

In this section, we introduce an image-to-image MLP-mixer architecture that builds on the original MLP-mixer~\citep{tolstikhin_MLPMixerAllMLPArchitecture_2021}. The image-to-image MLP-mixer operates on linearly transformed image patches, just like the MLP-mixer, as illustrated in Figure~\ref{fig:mixer}. 
However, contrary to the MLP-mixer, the image-to-image mixer imposes some structure by retaining the spacial order of image patches, which turns out to be critical for image reconstruction performance.

We start by splitting the image into non-overlapping patches of size $P\times P \times 3$ (our default choice is $P=4$). 

Each patch is viewed as a vector of dimension $3 P^2$ that is linearly transformed with the same trainable matrix to a space of arbitrary embedding dimension $C$. 
This patch embedding step thus transforms an image of dimension $H\times W \times 3$ (or $H\times W \times 1$ for greyscale images) to a volume of dimension ${H/P\times W/P \times C}$. The patch embedding step retains the relative positions of the patches in the image. The MLP-mixer and the vision transformer~\citep{tolstikhin_MLPMixerAllMLPArchitecture_2021, dosovitskiy2021an} also split an image into patches and linearly project the patches, and so do several other architectures for example the swin transformer~\citep{liu_SwinTransformerHierarchical_2021}.

We then apply an MLP-mixer layer inspired by 
the original MLP-mixer module. This MLP-mixer layer mixes the tensor in height dimension, then in width dimension, and finally in channel dimension. 
Mixing in channel dimension means viewing the tensor of dimension ${H/P\times W/P \times C}$ as a collection of $H/P \cdot W/P$ vectors of dimension $C$ and passing each of them through the same MLP consisting of a linear layer, followed by a GeLU non-linearity and then another linear layer. 
The hidden layer dimension is the input dimension of the respective vector multiplied by a factor of $f$. 
We also add skip connections and layer norms to help with the optimization. A mixer layer does not alter the dimensions of the input volume.

After $N$ many such mixer layers, the volume is transformed back to an image via a patch expansion step. The patch expansion step transforms the volume consisting of flattened patches, each of dimension $C$, back to an image of dimension $H/P\times W/P \times 3$ as follows: 
First, we linearly transform each patch of dimension $C$ to a patch of dimension $CP^2$ using a shared linear transformation.
This maps the volume of shape $H/P \times W/P \times C$ to a volume of shape $H/P \times W/P \times C P^2$. Second, we reshape the volume to a volume of shape $H\times W \times C$, and finally transform this volume to an image of shape $H\times W \times 3$ by linearly combining the layers (which can be implemented with a $1\times 1$ convolution).
A similar patch expansion step has been used by the Swin U-Net Transformer \citep{cao_SwinUnetUnetlikePure_2021}.

The main difference between our image-to-image MLP-mixer architecture and the original MLP-mixer is that we transform the image to a 3D tensor instead of a 2D tensor, and the mixer layer is modified to act on a 3D volume. This modification retains the relative location of the patches in the 3D volume which induces an inductive bias enabling the image-to-image MLP-mixer to perform very well when trained on relatively few images. 
As we show later in Section~\ref{sec:inductivebias}, the inductive bias is less than that of a convolutional network, but more than the original MLP-mixer.

A further difference of the image-to-image-Mixer over the original MLP-mixer is the scaling of the number of parameters:
The trainable parameters of the token mixing in the original mixer are $\mathcal{O}\left(H^2 W^2\right)$, while the height- and width mixing of the image-to-image MLP-mixer are $\mathcal{O}\left(H^2 + W^2 \right)$. The linear scaling in image resolution of the image-to-image MLP-mixer keeps the total number of trainable parameters low and the architecture memory efficient.  


\section{Experiments}

We evaluate the performance of the image-to-image mixer for a variety of image reconstruction problems. 
We focus on image denoising as it 
is considered to be a fundamental image reconstruction problem, for its practical importance, and since a good denoiser typically serves as a building block for other tasks such as recovering images from few and noisy measurements. 
For example, a state-of-the-art approach for reconstructing an image from few and noisy linear measurements is a so-called variational network which uses a denoiser as a building block~\citep{sriram_EndtoEndVariationalNetworks_2020}. Reconstructing an image from few and noisy linear measurements is an important inverse problem that arises in accelerated magnetic resonance imaging and sparse-view computed tomography. 

\paragraph{Baseline methods:}
We compare the denoising performance of the image-to-image MLP mixer to several baselines:

BM3D~\citep{dabov_ImageDenoisingSparse_2007}, a standard and well performing denoising algorithm that does not rely on any training data. 

The U-Net~\citep{ronneberger_UNetConvolutionalNetworks_2015}, a standard image-to-image convolutional network that is a go-to for image reconstruction problems. The U-Net performs slightly better than a standard multi-layer convolutional network for image denoising~\citep{brooks_UnprocessingImagesLearned_2019} (for example better than the DnCNN, a famous multi-layer convolutional network proposed by~\citet{zhang_GaussianDenoiserResidual_2017}). 

We also compare to the vision transformer~\citep{dosovitskiy2021an}, which we adapted for image recovery tasks as follows. We disposed the classification token and replaced the classification head by a linear layer that maps each element of the transformer output to a corresponding image patch. More details on how we adapted the ViT for recovery tasks are in the appendix. 

We finally  compare to the original MLP mixer, which we modified to perform image reconstruction as follows. We omitted the global average pooling and fully connected layer at the end, and used a projection matrix to linearly transform the hidden dimension $C$ back to dimension 
$3 P^2$. That results in a volume of dimension $S \times D$, where $S = HW/P^2$ and $D = 3 P^2$. Each row of dimension $D$ of the volume represents a flattened image patch. By unflattening each row in the table, i.e., by reshaping each row to dimensions $P \times P \times 3$, we end up with an image of dimension $H \times W \times 3$.

All networks (the image-to-image MLP-mixer, U-Net, ViT, and original MLP-mixer) are trained in the same fashion as described next.


\subsection{Gaussian Denoising}
\label{sec:gaussiandenoising}

We first consider the problem of removing Gaussian noise from ImageNet color images~\citep{deng_ImagenetLargescaleHierarchical_2009}.
We constructed a dataset as follows: We collected images of different classes from ImageNet and center-cropped them to a size of $256\times 256 \times 3$. We then added zero-mean Gaussian noise of standard deviation $\sigma=30$ to each image channel independently, resulting in a data set consisting of pairs of noisy image $\vy_i = \vx_i+\vz_i$ and corresponding clean image $\vx_i$. Here, $\vz_i$ is the Gaussian noise. The noisy images have a peak signal-to-noise-ratio (PSNR) of 19 dB. 

We trained the image-to-image MLP-mixer $f_\vtheta$ with trainable parameters $\vtheta$ (and the baseline architectures)
to map the noisy image to the noise by minimizing the loss function
\begin{align*}
\mc L(\vtheta) 
= 
\frac{1}{n} \sum_{i=1}^n
\frac{1}{2} \norm[2]{ \vy_i - f_{\vtheta}(\vy_i) - \vx_i}^2.
\end{align*}
Here, $n$ is the total number of training images. 
At inference, we are given a noisy image $\vy$ and estimate a clean image by subtracting the estimated residual from the noisy observation
as $\hat \vx = \vy - f_{\vtheta}(\vy)$. 
This is referred to as residual learning~\citep{zhang_GaussianDenoiserResidual_2017}, because the network learns to predict the residual. Training the network directly to map a noisy image to a clean image also works, but performs worse than residual learning for all architectures considered here.
 
We split the data set into train and test sets and ensured that images from the same ImageNet class do not exist in both sets simultaneously. This guarantees that the network is not just learning to denoise a specific class only.

We trained the different architectures on 100k images from the ImageNet training set. All networks have about 24M parameters. In Table \ref{tab:main_results}, we report the denoising results on our ImageNet test set, and also several other standard benchmarks: BSD68~\citep{BSD68}, Urban100~\citep{Urban100}, Kodak24\footnote{\url{http://r0k.us/graphics/kodak/}}, and McMaster18~\citep{McMaster}.

\begin{table}[h]
\centering
\small
\begin{tabular}{l c c c c c }
\toprule
 & ImageNet & BSD68 & Urban100 & Kodak24 & McMaster18\\ 
\midrule 
Img2Img-Mixer& $\bm{30.86}$& $\bm{30.15}$& $\bm{30.54}$& $\bm{31.37}$& $\bm{31.77}$\\ 
U-Net & 30.46 &29.88 &29.60 &30.95 &31.19\\ 
ViT &30.01 &29.57 &28.79 &30.51 &30.54 \\ 
Original-Mixer& 29.71& 29.21& 28.44& 30.08 &30.23\\ 
BM3D & 27.27 &26.89 &27.60 &28.11 &28.13 \\
\bottomrule
\end{tabular}
\caption{\label{tab:main_results} PSNR of the different networks for gaussian denoising. Our image-to-image Mixer outperforms all other networks on all datasets. The image-to-image Mixer also achieves the highest SSIM scores, as shown in table \ref{tab:main_results_ssim} in the appendix.  }
\end{table}

\paragraph{Scaling effect:}
The results reported in Table \ref{tab:main_results} were for large networks (24M) trained on a large dataset (100k). However, such large datasets are not always available, and often in memory limited applications, smaller models must be used. It is therefore interesting to study how the performance of the different networks changes when scaling the model and training set to smaller sizes. 

\begin{figure}[t]
\begin{center}
\begin{tikzpicture}

\begin{groupplot}[
cycle list name=Set1-5,
legend style={at={(1.35,1)},anchor=north}, 
legend style={cells={align=center}},
y label style={at={(axis description cs:0.20,0.5)},anchor=south},
group style={group size=3 by 1, horizontal sep=0.3cm, 
yticklabels at=edge left,
xticklabels at=edge bottom,
},
width=0.34\textwidth,height=0.31\textwidth, 
scaled x ticks=true,
every x tick label/.append style={alias=XTick,inner xsep=0pt},
every x tick scale label/.style={at=(XTick.base east),anchor=base west},
]

\nextgroupplot[title={4000 images},xlabel = {parameters (million)}, ylabel={PSNR (dB)},xmode=log,log basis x={2}, ymax=31.5, ymin=25]
\addplot[mark=*,color=blue] coordinates {(1.66,29.19) (2.4,29.56) (3.44,30.07) (6.61,30.20) (12.19, 30.22)};
\addplot[mark=x,color=orange] coordinates {(1.7,29.21) (2.45,29.45) (3.34,29.57) (6.82,29.87) (12.12, 29.92) };
\addplot[mark=diamond,color=magenta] coordinates {(1.69,23.68) (2.38,25.22) (3.68,26.35) (6.39,29.14) (12.19,29.17)};
\addplot[mark=square*,color=purple] coordinates {(1.69, 22.19)(2.42, 23.87) (3.36, 26.80) (6.87, 27.36) (12.21,27.36)};
\addplot[mark=none, color=teal] coordinates {(1.66,27.27) (12.19,27.27)};

\nextgroupplot[xshift=0.33cm, title={3M parameters},xlabel = {training images}, xmode=log,ymax=31.5, ymin=25]
\addplot[mark=*,color=blue] coordinates {(1000,28.13) (4000,30.07) (10000,30.32) (100000, 30.66 )};
\addplot[mark=x,color=orange] coordinates {(1000,28.12) (4000,29.57) (10000,29.76) (100000, 30.12)};
\addplot[mark=diamond,color=magenta] coordinates {(1000,26.12) (4000,26.35) (10000,26.40) (100000, 26.50)};
\addplot[mark=square*,color=purple] coordinates {(1000,24.85) (4000,26.80) (10000,26.97) (100000, 27.11)};
\addplot[mark=none, color=teal] coordinates {(1000,27.27)(100000,27.27)};

\nextgroupplot[title={24M parameters}, xlabel = {training images},xmode=log,ymax=31.5, ymin=25]
\addplot[mark=*,color=blue] coordinates { (1000,29.28) (4000,30.22) (10000,30.44) (100000, 30.86 )};
\addlegendentry{{\small Img2Img\\Mixer}}
\addplot[mark=x,color=orange] coordinates {(1000,29.62) (4000,30.05) (10000,30.24) (100000, 30.46)};
\addlegendentry{{\small U-Net}}
\addplot[mark=diamond,color=magenta] coordinates {(1000,28.17) (4000,29.17) (10000,29.87) (100000, 30.01)};
\addlegendentry{{\small ViT}}
\addplot[mark=square*,color=purple] coordinates {(1000,20.31) (4000,27.56) (10000,28.82) (100000, 29.71)};
\addlegendentry{{\small Original\\Mixer}}
\addplot[mark=none, color=teal] coordinates {(1000,27.27) (100000,27.27)};
\addlegendentry{{\small BM3D}}

\end{groupplot}

\end{tikzpicture}
\end{center}
\vspace{-0.3cm}
\caption{
\label{fig:example}
ImageNet Gaussian denoising performance for different networks when scaling the model and training set size. The PSNR of the noisy images is 19 dB.
{\bf Left:} Models of different sizes trained on 4000 images. The mixer outperforms all baselines for almost all sizes considered.
{\bf Middle:} Performance of networks with 3M parameters trained on a varying number of training images. For a very small number of training images, the U-Net performs on par with the image-to-image mixer, but in the regime where a moderate number of training examples is available, the mixer slightly outperforms the U-Net.
{\bf Right:} Same as the middle plot, but with a larger network size. It can be seen, that the slope of the U-Net starts reaching a plateau as the number of training images increases, but the image-to-image mixer continues to show an improvement in performance. 
}
\end{figure}
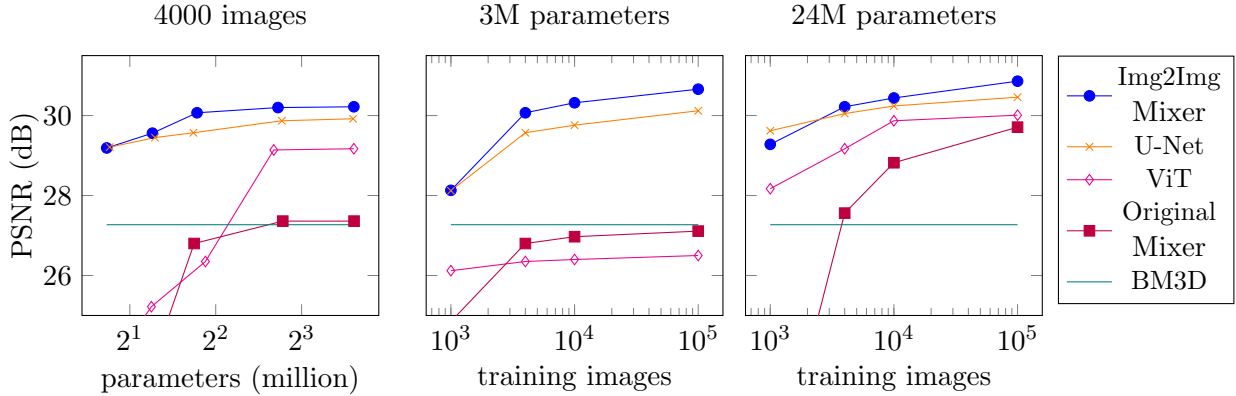

In Figure~\ref{fig:example}, we depict the denoising performance of the different architectures as a function of the number of training examples, ranging from 1000 to 100k training images, with constant model size, and as a function of the number of parameters, with constant training set size. The plots show that the ViT and the original mixer can only reach competitive performance when trained with a large model and training set. We hypothesize that this is due to their low inductive bias, which we measure in section \ref{sec:inductivebias}.

However, perhaps surprisingly, even in the regime of small training data ({\bf left:} 4000 images) and small model size ({\bf middle:} 3 million parameters), the image-to-image mixer can outperform the U-Net. On the left panel, it can be seen that the image-to-image MLP-Mixer is more parameter effective in that it reaches peak performance already at 3M parameters. It also outperforms the U-Net with fewer parameters, i.e., a 3M version of the image-to-image mixer performs slightly better than a 12M version of the U-Net. 

Most importantly, Figure~\ref{fig:example} shows that the image-to-image mixer scales better than the U-Net when both the dataset size and the size of the models grow.  
Particularly, the right panel shows that for large models (24M)
the gap in performance between the image-to-image mixer and the U-Net increases as the training set increases: 
 The U-Net shows a relatively smaller accuracy improvement when increasing the training set size from 10k to 100k. Thus, we expect even larger improvements when moving to even larger datasets.

In the experiment, the model parameters of the original and the image-to-image mixer are varied by changing the number of layers, embedding dimension, and hidden dimension of the MLPs. The exact hyperparameter configurations of the image-to-image mixer are in Table \ref{tab:param_config} in the appendix. For the U-Net, we increased the model size by increasing the number of channels, and for ViT we increased the model size by increasing both its depth and width. 
 
\subsection{Denoising performance on real-world camera noise}

We next evaluate the performance of the image-to-image MLP-mixer on real-world camera noise, which is often not well approximated by Gaussian noise. 
We evaluate on the Smartphone Image Denoising Dataset (SIDD)~\citep{abdelhamed_HighQualityDenoisingDataset_2018}, which consists of high-resolution images from 10 scenes obtained under different lighting conditions with five representative smartphone cameras. 
We center-cropped a $2048\times 2048$ patch from each image and divided that into non-overlapping images of size $256\times 256$. We used 4700 images from 8 of the scenes for training and 700 images from the remaining 2 scenes for testing.

Unlike the Gaussian denoising setup studied before, each SIDD image has a different noise level.  The noisy images have an average PSNR of 24 dB, 5 dB higher than the noisy ImageNet images, thus most SIDD images have a much lower noise level than the ImageNet images we denoised in the previous section. 
The image-to-image mixer achieved a denoising PSNR of 33.66 dB, whereas the U-Net, the ViT, and the original mixer obtained lower values of 33.13 dB and 32.87 dB, and 29.49 dB respectively. All networks had about 7M parameters. BM3D requires one hyperparameter for the noise variance. We estimated the noise level from the ground truths and used it as an input to BM3D, which achieved 28.87 dB (the ground truth it typically not available, but this results in the best possible performance). 

This experiment shows that the good performance of the image-to-image MLP-mixer is not only limited to Gaussian denoising or a specific noise variance: The image-to-image mixer also outperformes the other baselines on real-world camera noise of different noise levels.

\subsection{Compressive sensing}

Next, we evaluate the image-to-image mixer on the task of recovering an image $\vx \in \reals^n$ from few linear measurements $\vy = \mA \vx$, where $\mA \in \reals^{m\times n}$, with $m < n$, is a wide and known measurement matrix. This compressive sensing problem arises in sparse-view tomography and accelerated magnetic resonance imaging. 
Our results show that, similar to the previous section, the performance of the image-to-image MLP-mixer scales well with the number of training images and size of the network.

A standard approach to address the compressive sensing problem with a neural network is to first compute a coarse reconstruction via least-squares as $\pinv{\mA} \vy$ and then train a neural network to map the coarse least-squares reconstruction to a clean reconstruction by minimizing the loss $\mc L(\vtheta)= \frac{1}{n}\sum_{i=1}^n \left(1 - 
\mathrm{SSIM}(f_\vtheta(\pinv{\mA}\vy_i), \vx_i) \right)$, where $f_\vtheta$ is a neural network with parameters $\vtheta$ mapping an image to an image. 
Here, SSIM is the structural similarity index metric~\citep{zhouwang_ImageQualityAssessment_2004}, a metric indicating the visual similarity between two images, larger is better, and a value of $1$ indicates that the two images are equivalent. 
This approach has been pioneered by~\citet{jin_DeepConvolutionalNeural_2017} for computational tomography and serves as a baseline for a competition for accelerated MRI, called FastMRI~\citep{zbontar_FastMRIOpenDataset_2018}.  

\begin{figure}[t]
\begin{center}
\begin{tikzpicture}
\begin{groupplot}[
cycle list name=Set1-5,
legend style={at={(1.4,1)},anchor=north}, 
legend style={cells={align=center}},
group style={group size=2 by 1, horizontal sep=1.5cm },
width=0.39\textwidth,height=0.31\textwidth, 
scaled x ticks=true,
every x tick label/.append style={alias=XTick,inner xsep=0pt},
every x tick scale label/.style={at=(XTick.base east),anchor=base west},
]
\nextgroupplot[xlabel = {number of parameters (millions)},ylabel={SSIM},xmode=log,log basis x={2},scaled y ticks=false,
yticklabel=\pgfkeys{/pgf/number format/.cd,fixed,precision=3,zerofill}\pgfmathprintnumber{\tick},]

\addplot[mark=*,color=blue] coordinates {(8,0.7422) (16,0.7442) (32,0.7447)};

\addplot[mark=x,color=orange] coordinates {(8,0.7423) (16,0.7430) (32,0.7432)};
\addplot[mark=diamond,color=magenta] coordinates {(8,0.7407) (16,0.7423) (32,0.7438)};


\nextgroupplot[xlabel = {number of training images}, xtick={2, 10, 17, 35}, xticklabels={2k, 10k, 17k, 35k}, scaled y ticks=false,
yticklabel=\pgfkeys{/pgf/number format/.cd,fixed,precision=2,zerofill}\pgfmathprintnumber{\tick},]

\addplot[mark=*,color=blue] coordinates {(2,0.7211) (10,0.7384) (17,0.7408) (35,0.7422)};
\addlegendentry{Img2Img\\Mixer}

\addplot[mark=x,color=orange] coordinates {(2,0.7360) (10,0.7405) (17,0.7414) (35,0.7423)};
\addlegendentry{U-Net}

\addplot[mark=diamond,color=magenta] coordinates {(2,0.7227) (10,0.7345) (17,0.7391) (35,0.7407)};
\addlegendentry{ViT}

\end{groupplot}          
\end{tikzpicture}
\end{center}
\vspace{-0.3cm}
\caption{
\label{fig:cs_accuracy}
\textbf{Left:} Compressed sensing performance for different model sizes when trained on the entire fastMRI knee dataset (35k data). Contrary to the image-to-image mixer, the U-Net's performance does not benefit from the increase in the number of parameters. \textbf{Right:} Compressed sensing performance as a function of the number of training images for networks of size about 8 million parameters. For a small number of training images, the U-Net reaches higher SSIM scores than the image-to-image MLP-mixer, but this performance gap quickly diminishes as training data grows. 
}
\end{figure}
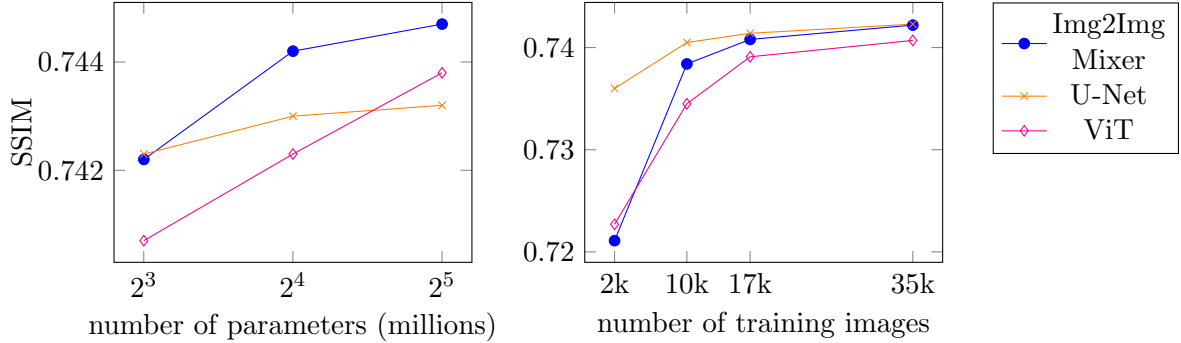

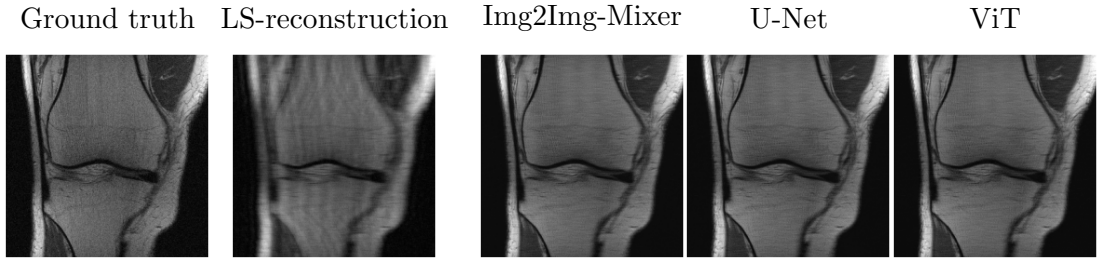
\begin{figure}[t]
\begin{center}

\resizebox{\textwidth}{!}{%

\begin{tikzpicture}

\newcommand\xxspace{2.7}
\newcommand\yspace{1}
\newcommand\ymargin{1}
\newcommand\xmargin{1}
\newcommand\ycap{-2.8cm}
\newcommand\iwidth{2.8cm}


\node at (0*\xxspace,-1*\yspace) {\includegraphics[width=\iwidth, angle=180]{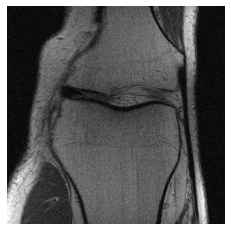}};
\node at (1.1*\xxspace,-1*\yspace) {\includegraphics[width=\iwidth, angle=180]{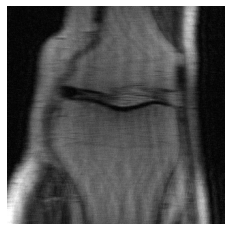}};
\node at (2.3*\xxspace,-1*\yspace) {\includegraphics[width=\iwidth, angle=180]{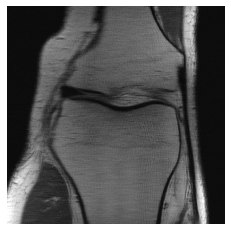}};
\node at (3.3*\xxspace,-1*\yspace) {\includegraphics[width=\iwidth, angle=180]{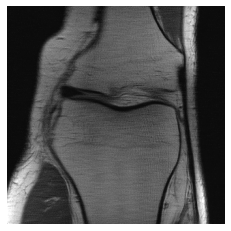}};
\node at (4.3*\xxspace,-1*\yspace) {\includegraphics[width=\iwidth, angle=180]{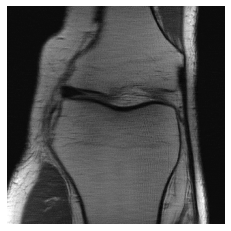}};

\node at (0*\xxspace,0.8*\yspace) {\parbox{4cm}{\centering Ground truth}};
\node at (1.1*\xxspace,0.8*\yspace) {\parbox{4cm}{\centering LS-reconstruction
}};
\node at (2.3*\xxspace,0.8*\yspace) {\parbox{4cm}{\centering Img2Img-Mixer
}};
\node at (3.3*\xxspace,0.8*\yspace) {\parbox{4cm}{\centering U-Net
}};
\node at (4.3*\xxspace,0.8*\yspace) {\parbox{4cm}{\centering ViT
}};

\end{tikzpicture}
}
\end{center}
\vspace{-0.4cm}
\caption{
\label{fig:exampleCS}
We trained the different architectures to map a coarse least-squares (LS)-reconstruction to the ground truth. 
All networks yield very similar reconstruction performance, three example images are given, for training on 35k images.
}
\end{figure}

We evaluate the image-to-image mixer, the U-Net, and the ViT on a four-times accelerated MRI knee-reconstruction problem (i.e., $m = n/4$). 
We trained the networks of equal size of about 8 million parameters on the FastMRI knee training dataset containing 
2k, 10k, 17k, and 35k training images and evaluated their performance on the FastMRI knee validation set. Figure~\ref{fig:cs_accuracy} depicts the reconstruction performance as a function of the number of training examples.
Example reconstructions are given in Figure~\ref{fig:exampleCS}. 
In this experiment, all three architectures yield similar performance. 
The same trends as in the denoising experiment in Figure~\ref{fig:example} hold true: 
that the performance of the image-to-image MLP-mixer scales well with number of training images and size of the network, and surpases that of U-Net if the model is sufficiently large and trained on sufficiently many images. 

The results show that the image-to-image mixer yields competitive performance beyond plain denoising tasks.
Together, our denoising and compressive sensing results demonstrate that convolutions and a multi-resolution architecture are not necessary for state-of-the-art imaging performance.




\subsection{Measuring the inductive bias of the Image-to-Image Mixer}
\label{sec:inductivebias}

In this section we measure the inductive bias of the different architectures considered here. We find that low inductive bias correlates with more significant performance improvements as both the model size and size of the dataset are increased: Both the original and image-to-image MLP-mixer, and the ViT have a lower inductive bias than the U-Net (as shown in this section) and therefore show a larger increase in performance as the number of parameters and dataset are increased (see Figure~\ref{fig:example}).

Convolutional neural networks have an inductive towards natural images in that they are well suited to generate natural images. The inductive bias of convolutional neural networks is so strong that a convolutional neural network can perform image reconstruction without any training. This has first been shown for a U-Net in the deep image prior paper: \citet{ulyanov_DeepImagePrior_2018} has shown that a randomly initialized, un-trained U-Net fits a natural image with significantly fewer gradient descent iterations than it fits noise. 
This effect can be reproduced with a very simple convolutional network, without any skip connections and without an encoder-decoder structure~\citep{heckel_DeepDecoderConcise_2019}. This inductive bias has been theoretically explained by wide convolutional networks trained with gradient descent fitting the lower frequencies of a signal before fitting the higher frequencies~\citep{heckel_DenoisingRegularizationExploiting_2020}. Since a natural image has much of its energy concentrated on low-frequency components, a natural image is fitted faster than Gaussian noise which has in expectation the same energy on all components.

\begin{figure}[t]
\begin{center}
\begin{tikzpicture}
\begin{groupplot}[
y tick label style={/pgf/number format/.cd,fixed,precision=3},
scaled y ticks = false, 
group style={group size= 4 by 2, xlabels at=edge bottom, horizontal sep=0.1cm, vertical sep=0.1cm,
yticklabels at=edge left,
xticklabels at=edge bottom,
},
xmin=5,
width=0.28*\textwidth,height=0.28*\textwidth]
\nextgroupplot[title={U-Net},ylabel={MSE}, ymax=0.15, ymin=0.008, xmode=log,ymode=log] 
\addplot +[mark=none,color=violet,line width=0.8pt] table[x index=0,y index=1]{./untrained/unet/unet_noisy_img.dat};

\nextgroupplot[title={Img2Img-Mixer}, 
xmode=log,  ymax=0.15, ymin=0.008,ymode=log] 
\addplot +[mark=none,color=violet,line width=0.8pt] table[x index=0,y index=1]{./untrained/mixer/mixer_noisy_img.dat};

\nextgroupplot[title={ViT}, ymax=0.15, ymin=0.008, xmode=log,ymode=log]
\addplot +[mark=none,color=violet,line width=0.8pt] table[x index=0,y index=1]{./untrained/vit/vit_noisy_img.dat};

\nextgroupplot[title={Original-Mixer}, ymax=0.15, ymin=0.008, xmode=log,ymode=log]
\addplot +[mark=none,color=violet,line width=0.8pt,] table[x index=0,y index=1]{./untrained/origin_mixer/origin_mixer_noisy_img.dat};
\addlegendentry{{\small img + noise}}


\nextgroupplot[ylabel={MSE}, xmode=log, ymax=0.2,ymin=0.0] 
\addplot +[mark=none,line width=0.8pt] table[x index=0,y index=1]{./untrained/unet/unet_clean.dat};
\addplot +[mark=none,line width=0.8pt] table[x index=0,y index=1]{./untrained/unet/unet_noise.dat};	

\nextgroupplot[xlabel={iteration}, xlabel style={xshift=0.45in}, xmode=log,ymax=0.2,ymin=0.0]
\addplot +[mark=none,line width=0.8pt] table[x index=0,y index=1]{./untrained/mixer/mixer_clean.dat};
\addplot +[mark=none,line width=0.8pt] table[x index=0,y index=1]{./untrained/mixer/mixer_noise.dat};

\nextgroupplot[xmode=log,ymax=0.2,ymin=0.0] 
\addplot +[mark=none,line width=0.8pt] table[x index=0,y index=1]{./untrained/vit/vit_clean.dat};
\addplot +[mark=none,line width=0.8pt] table[x index=0,y index=1]{./untrained/vit/vit_noise.dat};	

\nextgroupplot[xmode=log,ymax=0.2,ymin=0.0] 
\addplot +[mark=none,line width=0.8pt] table[x index=0,y index=1]{./untrained/origin_mixer/origin_mixer_clean.dat};
\addlegendentry{{\small img}}
\addplot +[mark=none,line width=0.8pt] table[x index=0,y index=1]{./untrained/origin_mixer/origin_mixer_noise.dat};
\addlegendentry{{\small noise}}

\end{groupplot}         

\node at (13.8,2.1) {\includegraphics[width=0.5in]{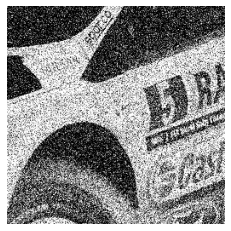}};
\node at (13.8,2.9) {img + noise}; 
\node at (13.8,-0.1) {\includegraphics[width=0.5in]{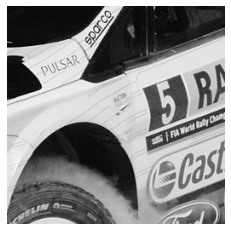}};
\node at (13.8,0.7) {img}; 
\node at (13.8,-2.3) {\includegraphics[width=0.5in]{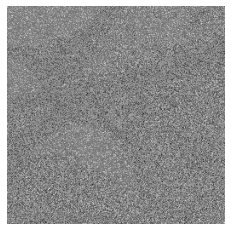}};\node at (13.8,-1.5) {noise}; 
\end{tikzpicture}
\end{center}
\vspace{-0.3cm}
\caption{
\label{fig:untrained}
Measuring the inductive bias of architectures. 
We fitted the \{Img2Img-Mixer,U-Net,ViT, Original-Mixer\} to map a random input to (a) an image, (b) Gaussian noise, and (c) the image plus noise. 
MSE denotes Mean Square Error of fitting the network output with respect to the clean image for img and img + noise, and with respect to the noise for noise. 
All networks have significantly more parameters than pixels and can fit both noise and a natural image perfectly, but have an inductive bias, in that they fit the image faster than the noise. 
}
\end{figure}
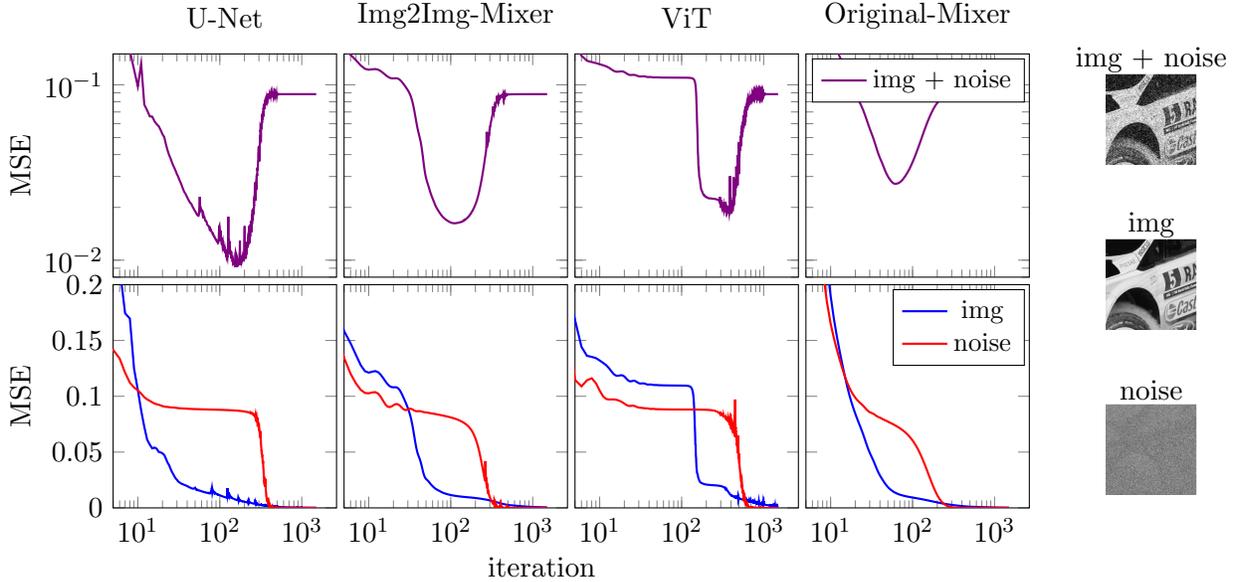

Motivated by this observation, we measured the inductive bias of the image-to-image mixer, the original MLP-mixer, the U-Net, and the ViT, by fitting the respective randomly initialized networks to i) a natural image, ii) Gaussian noise, and iii) the natural image plus the Gaussian noise. The three signals are displayed in Figure~\ref{fig:untrained} along with the training curves obtained by minimizing the loss 
$\mc L(\vtheta) = \frac{1}{2}\norm[2]{\text{signal} - f_{\vtheta}(\vy)}$ with Gradient descent. Here, signal is the respective signal (i.e., img, noise, and img+noise), and $f_\vtheta(\vy)$ is the respective network, initialized randomly, and fed with a random input $\vy$.

Figure~\ref{fig:untrained} shows that the U-Net has a larger inductive bias than the image-to-image mixer, which in turn has a larger inductive bias than the ViT and the original MLP-mixer: without any training, the U-Net achieves a denoising performance of 20.2 dB, the image-to-image mixer of 18.1 dB, the ViT of 17.3 dB and the original MLP-mixer of 16.3 dB. The noisy image in that case is 12.5 dB (see Figure~\ref{fig:examplesuntrained}).
Interestingly, all four networks have an inductive bias in that they fit a natural image significantly faster than noise. 

Figure~\ref{fig:examplesuntrained} in the appendix illustrates the type of inductive bias: The convolutional U-Net has an inductive bias towards smooth signals, the image-to-image mixer towards fitting vertical and horizontal lines first.

\subsection{Discussion}

We saw in Section \ref{sec:inductivebias} that the inductive bias of the image-to-image mixer lies in between that of the U-Net and the original mixer. The low inductive bias of the original mixer results in very poor performance when the sizes of the training set and model are small, as shown in Figure \ref{fig:example}. The U-Net's high inductive bias enables it to perform very well when trained on few images compared to the other networks (see right panel in Figure \ref{fig:cs_accuracy}), but limits its scalibility in the regime of large models and datasets, as seen in the right panel of Figure \ref{fig:example} and the left panel of Figure \ref{fig:cs_accuracy}. The image-to-image mixer serves as a middle ground, where it has a moderate inductive bias that allows it to perform very well in the regime of small model and training set, but also permits it to scale well when increasing the training set and model sizes.

The poor performance of the original mixer compared to the image-to-image mixer highlights the merits of the modifications we introduced in this paper. Retaining the relative position of the image patches during the patch embedding operations is critical for inducing a bias towards natural images. Mixing separately in the height and width dimensions as opposed to token (height and width dimensions merged together) mixing as in the original mixer not only reduces the number of trainable parameters, but also boosts performance significantly. Unfortunately, there is a downside to our modifications, namely that they reduce the speed of the network. The good performance of the image-to-image mixer comes at the cost of a lower throughput compared to the original mixer, as shown later in table \ref{tab:throughput}.

\subsection{Ablation studies}
\label{sec:ablation}

In this section we discuss a few variants of the image-to-image mixer in order to understand which elements of the network are critical for its image reconstruction performance. Further ablation studies can be found in the appendix.

\paragraph{Incorporating multi-resolution.} The most successful architectures to date for image reconstruction and dense predictions incorporate a notion of multi-scale. For example, the U-Net~\citep{ronneberger_UNetConvolutionalNetworks_2015} transforms an image by first decreasing the spacial dimensions and increasing number of channels, and second increasing the spacial dimensions while decreasing the number of channels.
Even image-to-image transformers (or attention-based networks)  incorporate such multi-resolution structure~\citep{liu_SwinTransformerHierarchical_2021} successfully. 
For convolutional architectures, incorporating such multi-scale architecture improves performance, in that the U-Net outperforms a standard multi-layer convolutional network at denoising~\citep{brooks_UnprocessingImagesLearned_2019}.

We incorporated such multi-resolution structures by implementing patch merging as in the Swin Transformer ~\citep{liu_SwinTransformerHierarchical_2021} and patch expanding as in the Swin U-Net Transformer~\citep{cao_SwinUnetUnetlikePure_2021}. Patch merging can be seen as an encoding step, where the spatial dimensions (height and width) are reduced by a factor of two and the channel dimension is increased by a factor of two. Patch expanding acts as the decoder by reversing the merging operation, i.e., it increases the spatial dimensions and decreases the number of channels. The merging and expanding steps are implemented by linear transformations and reshaping as in the patch combining step. Figure 1 in the paper~\citep{cao_SwinUnetUnetlikePure_2021} visualizes the similar Swin Transformer architecture, but instead of swin transformer blocks we used the mixer layers.

Our results show that incorporating multi-resolution structure does not improve performance and instead marginally decreases performance. 
We considered a multi-resolution image-to-image mixer and compared it to our proposed image-to-image mixer, and the other baselines on the Gaussian denoising experiment described in Section~\ref{sec:gaussiandenoising} (7M parameters and 4000 training images). 
The multi-resolution mixer achieved 28.77 dB, less than the Img2Img-Mixer (30.20 dB), the U-Net (29.87 dB), and the ViT (29.14 dB), but more than the original mixer (27.36 dB). We also evaluated the multi-resolution architecture on the SIDD images, where it achieved 33.48 dB, better than the U-Net (33.13 dB), the ViT (32.87 dB), and the original mixer (29.49 dB) but still slightly worse than the Img2Img-Mixer (33.66 dB). However, incorporating a multi-resolution structure significantly improves the throughput when using large batch sizes, as shown later.

\paragraph{Throughput}
We measured the throughput of the networks by calculating the average speed of a forward pass on the GPU at inference. Since the networks benefit from different batch sizes, we report the results in Table \ref{tab:throughput} for the best performing batch size and also for batch size = 1, which is most relevant if we process one image at inference, which is common for example in the MRI application we discussed. All networks have a size of about 3M parameters. In addition to the baselines, we report the throughput for the multi-resolution image-to-image mixer discussed earlier, and also for DnCNN~\citep{zhang_GaussianDenoiserResidual_2017}, a popular network for denoising.  
For batch size 1, the image-to-image mixer is faster than the DnCNN and similar to ViT and the multi-resolution mixer, but slower than the U-Net and the original mixer.

\begin{table}[h]
\centering
\small
\begin{tabular}{l c c c c c c }
\toprule
Batch Size & Original-Mixer & ViT & U-Net & Multi-Resolution-Mixer & Img2Img-Mixer & DnCNN\\ 
\midrule 
1 & 205 & 116 & 287 & 97 & 89 & 28 \\ 
best & 925 & 690 & 524 & 442 &98 & 30 \\ 

\bottomrule

\end{tabular}

\caption{\label{tab:throughput} Throughput of the different networks measured at inference as number of images per second per GPU.}
\end{table}

\section{Related literature}

Our work builds on the recently introduced MLP-mixer~\citep{tolstikhin_MLPMixerAllMLPArchitecture_2021}. While there are a number of works that also build on the MLP-mixer, to the best of our knowledge, this is the first work exploring a structured MLP-based architecture for image reconstruction tasks. 

There are several recent works that build on the MLP-mixer for classification tasks: \citet{chen_CycleMLPMLPlikeArchitecture_2021} proposed to mix the spatial dimensions in a cyclic way resulting in an architecture that performs well on detection and segmentation tasks. 
The ResMLP network~\citep{touvron_ResMLPFeedforwardNetworks_2021} replaces the self-attention layers of a ViT by an MLP, yielding competitive image classification performance. RaftMLP~\citep{tatsunami_RaftMLPMLPbasedModels_2021} modifies the MLP mixer for classification by mixing the spacial dimensions in a similar way as we do, and achieve a more parameter efficient model for classification.
\citet{cazenavette_MixerGANMLPBasedArchitecture_2021} proposed an image-to-image GAN that utilizes MLP-mixer blocks followed by convolutional layers in the decoder part. 
\citet{liu_PayAttentionMLPs_2021a} proposed to substitute attention with MLPs paired with gating, demonstrating that attention is not critical for ViTs to perform well.

We finally note that even a completely unstrutured MLP can perform well for denoising small image patches. Specifically~\citet{burger_ImageDenoisingCan_2012} trained an MLP to denoise image patches of size $17\times 17$ and achieved performance competitive with BM3D.

\section{Conclusion}

We introduced and evaluated a simple architecture based on the MLP-mixer~\citep{tolstikhin_MLPMixerAllMLPArchitecture_2021} for image-to-image reconstruction tasks. Image reconstruction tasks are currently dominated by convolutional networks that incorporate a multi-resolution structure such as the U-Net. Our work shows that an architecture based on MLPs and without a multi-resolution structure gives even slightly better performance at both small, moderate, and large network sizes for denoising. Unlike the original mixer, the image-to-image architecture incorporates structure by retaining the relative positions of the image patches. This induces a bias towards natural images, that is higher than the original mixer and the ViT, but lower than U-Net, which our experiments show to give the best trade off. 

If trained on a moderate amount of images, the image-to-image MLP mixer slightly outperforms both the U-Net as well as the ViT at synthetic denoising tasks and at real-world denoising tasks. For compressive sensing, we found all architectures to perform very similarly. 
Our work shows that training on millions of images is not essential for non convolutional networks to compete with CNNs. Even in the regime of moderately sized training sets, CNNs can be outperformed. The necessity of massive datasets has been a limiting factor for further research in non-convolutional networks. We therefore hope that our work serves as a starting point for ending the dominance of CNNs in image processing tasks, and motivates further research in models free from hand-crafted visual features and high inductive biases.  


\section{Reproducibility statement}

The code to reproduce the results in this paper is available on our Github page: \url{https://github.com/MLI-lab/imaging_MLPs}. The experiments were carried out on a server with four RTX6000 GPUs, most experiments reported here run on a single GPU for less than a day.

{ 
\AtNextBibliography{\small} 
\printbibliography
}

\appendix

\section{Supplementary material}

\begin{figure}[H]
\begin{center}

\resizebox{\textwidth}{!}{%

\begin{tikzpicture}

\newcommand\xxspace{2.7}
\newcommand\yspace{1}
\newcommand\ymargin{1}
\newcommand\xmargin{1}
\newcommand\ycap{-2.8cm}
\newcommand\iwidth{2.8cm}

\node[align=left] at (-1*\xxspace,-1*\yspace) {Img2Img-\\Mixer};
\node at (0*\xxspace,-1*\yspace) {\includegraphics[width=\iwidth]{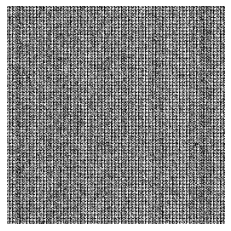}};
\node at (1*\xxspace,-1*\yspace) {\includegraphics[width=\iwidth]{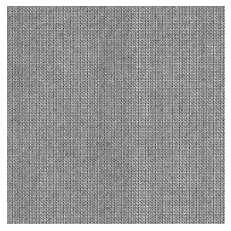}};
\node at (2*\xxspace,-1*\yspace) {\includegraphics[width=\iwidth]{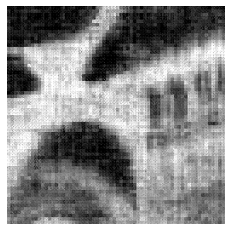}};
\node at (3*\xxspace,-1*\yspace) {\includegraphics[width=\iwidth]{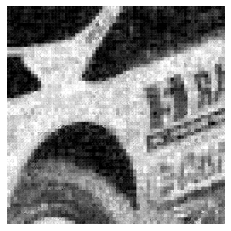}};
\node at (4*\xxspace,-1*\yspace) {\includegraphics[width=\iwidth]{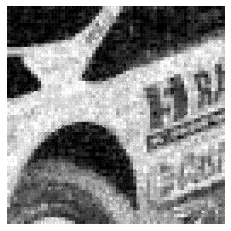}};
\node at (5*\xxspace,-1*\yspace) {\includegraphics[width=\iwidth]{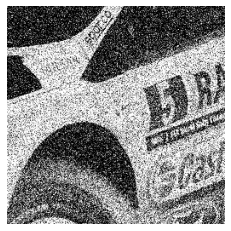}};
\node at (6*\xxspace,-1*\yspace) {\includegraphics[width=\iwidth]{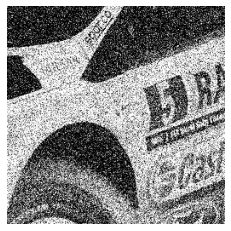}};
\node at (7*\xxspace,-1*\yspace) {\includegraphics[width=\iwidth]{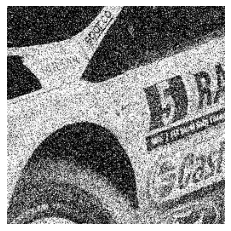}};

\node at (0*\xxspace,0.8*\yspace) {\parbox{4cm}{\centering Iter.: 1\\  6.6 dB}};
\node at (1*\xxspace,0.8*\yspace) {\parbox{4cm}{\centering Iter.: 10\\ 9.1 dB}};
\node at (2*\xxspace,0.8*\yspace) {\parbox{4cm}{\centering Iter.: 50\\  15.7 dB}};
\node at (3*\xxspace,0.8*\yspace) {\parbox{4cm}{\centering Iter.: 100\\ 18.1 dB}};
\node at (4*\xxspace,0.8*\yspace) {\parbox{4cm}{\centering \textcolor{red}{Best Iter.: 111\\ 18.1 dB}}};
\node at (5*\xxspace,0.8*\yspace) {\parbox{4cm}{\centering Iter.: 700\\  12.5 dB}};
\node at (6*\xxspace,0.8*\yspace) {\parbox{4cm}{\centering Iter.: 1100\\  12.5 dB}};
\node at (7*\xxspace,0.8*\yspace) {\parbox{4cm}{\centering Iter.: 1500\\  12.5 dB}};

\node at (8.2*\xxspace,-1*\yspace) {\includegraphics[width=\iwidth]{./untrained/noisy_img.png}};
\node at (9.2*\xxspace,-1*\yspace) {\includegraphics[width=\iwidth]{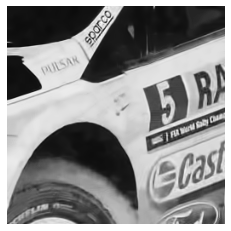}};
\node at (10.2*\xxspace,-1*\yspace) {\includegraphics[width=\iwidth]{./untrained/clean.png}};

\node at (8.2*\xxspace,0.8*\yspace) {\parbox{4cm}{\centering Noisy Image\\  12.5 dB}};
\node at (9.2*\xxspace,0.8*\yspace) {\parbox{4cm}{\centering BM3D\\ 23.5 dB}};
\node at (10.2*\xxspace,0.8*\yspace) {\parbox{4cm}{\centering Clean Image}};

\node[align=left] at (-1*\xxspace,-5*\yspace) {U-Net};
\node at (0*\xxspace,-5*\yspace) {\includegraphics[width=\iwidth]{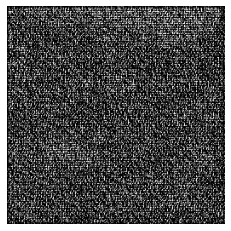}};
\node at (1*\xxspace,-5*\yspace) {\includegraphics[width=\iwidth]{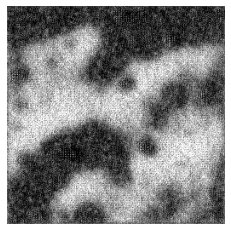}};
\node at (2*\xxspace,-5*\yspace) {\includegraphics[width=\iwidth]{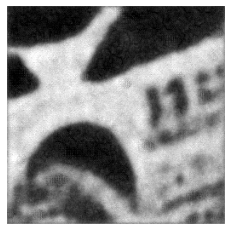}};
\node at (3*\xxspace,-5*\yspace) {\includegraphics[width=\iwidth]{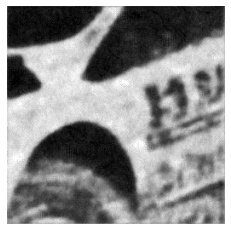}};
\node at (4*\xxspace,-5*\yspace) {\includegraphics[width=\iwidth]{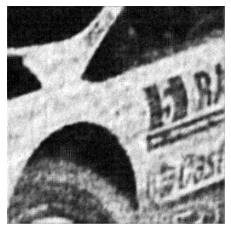}};
\node at (5*\xxspace,-5*\yspace) {\includegraphics[width=\iwidth]{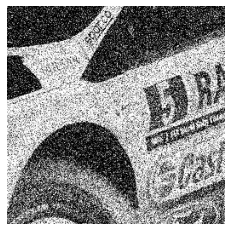}};
\node at (6*\xxspace,-5*\yspace) {\includegraphics[width=\iwidth]{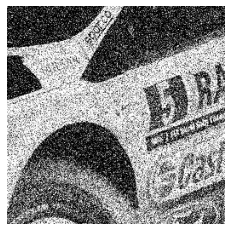}};
\node at (7*\xxspace,-5*\yspace) {\includegraphics[width=\iwidth]{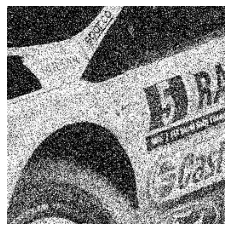}};

\node at (0*\xxspace,-3.2*\yspace) {\parbox{4cm}{\centering Iter.: 1\\ 5.4 dB}};
\node at (1*\xxspace,-3.2*\yspace) {\parbox{4cm}{\centering Iter.: 10\\  11.2 dB}};
\node at (2*\xxspace,-3.2*\yspace) {\parbox{4cm}{\centering Iter.: 50\\  17.4 dB}};
\node at (3*\xxspace,-3.2*\yspace) {\parbox{4cm}{\centering Iter.: 100\\  18.3 dB}};
\node at (4*\xxspace,-3.2*\yspace) {\parbox{4cm}{\centering \textcolor{red}{Best Iter.: 157\\  20.2 dB}}};
\node at (5*\xxspace,-3.2*\yspace) {\parbox{4cm}{\centering Iter.: 700\\ 12.5 dB}};
\node at (6*\xxspace,-3.2*\yspace) {\parbox{4cm}{\centering Iter.: 1100\\ 12.5 dB}};
\node at (7*\xxspace,-3.2*\yspace) {\parbox{4cm}{\centering Iter.: 1500\\ 12.5 dB}};

\node at (8.2*\xxspace,-5*\yspace) {\includegraphics[width=\iwidth]{./untrained/noisy_img.png}};
\node at (9.2*\xxspace,-5*\yspace) {\includegraphics[width=\iwidth]{./untrained/bm3d.png}};
\node at (10.2*\xxspace,-5*\yspace) {\includegraphics[width=\iwidth]{./untrained/clean.png}};

\node at (8.2*\xxspace,-3.2*\yspace) {\parbox{4cm}{\centering Noisy Image\\  12.5 dB}};
\node at (9.2*\xxspace,-3.2*\yspace) {\parbox{4cm}{\centering BM3D\\ 23.5 dB}};
\node at (10.2*\xxspace,-3.2*\yspace) {\parbox{4cm}{\centering Clean Image}};

\node[align=left] at (-1*\xxspace,-9*\yspace) {ViT};
\node at (0*\xxspace,-9*\yspace) {\includegraphics[width=\iwidth]{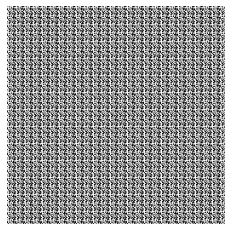}};
\node at (1*\xxspace,-9*\yspace) {\includegraphics[width=\iwidth]{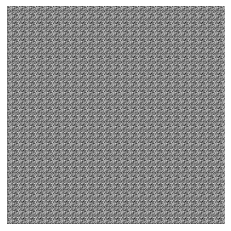}};
\node at (2*\xxspace,-9*\yspace) {\includegraphics[width=\iwidth]{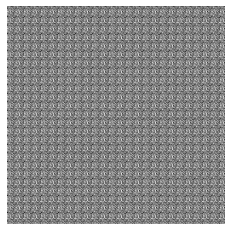}};
\node at (3*\xxspace,-9*\yspace) {\includegraphics[width=\iwidth]{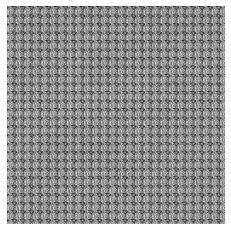}};
\node at (4*\xxspace,-9*\yspace) {\includegraphics[width=\iwidth]{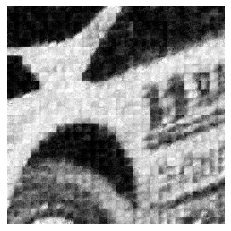}};
\node at (5*\xxspace,-9*\yspace) {\includegraphics[width=\iwidth]{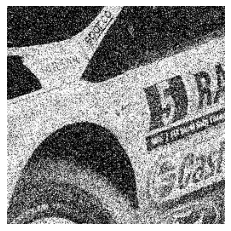}};
\node at (6*\xxspace,-9*\yspace) {\includegraphics[width=\iwidth]{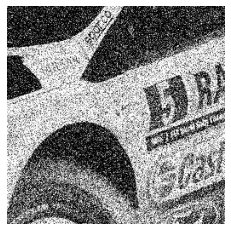}};
\node at (7*\xxspace,-9*\yspace) {\includegraphics[width=\iwidth]{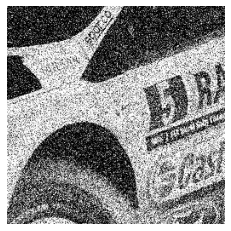}};

\node at (0*\xxspace,-7.2*\yspace) {\parbox{4cm}{\centering Iter.: 1\\  5.7 dB}};
\node at (1*\xxspace,-7.2*\yspace) {\parbox{4cm}{\centering Iter.: 10\\  9.1 dB}};
\node at (2*\xxspace,-7.2*\yspace) {\parbox{4cm}{\centering Iter.: 50\\ 9.6 dB}};
\node at (3*\xxspace,-7.2*\yspace) {\parbox{4cm}{\centering Iter.: 100\\  9.6 dB}};
\node at (4*\xxspace,-7.2*\yspace) {\parbox{4cm}{\centering \textcolor{red}{Best Iter.: 363\\ 17.3 dB}}};
\node at (5*\xxspace,-7.2*\yspace) {\parbox{4cm}{\centering Iter.: 700\\  11.9 dB}};
\node at (6*\xxspace,-7.2*\yspace) {\parbox{4cm}{\centering Iter.: 1100\\  11.9 dB}};
\node at (7*\xxspace,-7.2*\yspace) {\parbox{4cm}{\centering Iter.: 1500\\ 11.8 dB}};

\node at (8.2*\xxspace,-9*\yspace) {\includegraphics[width=\iwidth]{./untrained/noisy_img.png}};
\node at (9.2*\xxspace,-9*\yspace) {\includegraphics[width=\iwidth]{./untrained/bm3d.png}};
\node at (10.2*\xxspace,-9*\yspace) {\includegraphics[width=\iwidth]{./untrained/clean.png}};

\node at (8.2*\xxspace,-7.2*\yspace) {\parbox{4cm}{\centering Noisy Image\\  12.5 dB}};
\node at (9.2*\xxspace,-7.2*\yspace) {\parbox{4cm}{\centering BM3D\\ 23.5 dB}};
\node at (10.2*\xxspace,-7.2*\yspace) {\parbox{4cm}{\centering Clean Image}};

\node[align=left] at (-1*\xxspace,-13*\yspace) {Original-\\Mixer};
\node at (0*\xxspace,-13*\yspace) {\includegraphics[width=\iwidth]{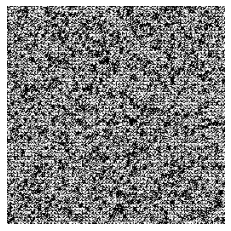}};
\node at (1*\xxspace,-13*\yspace) {\includegraphics[width=\iwidth]{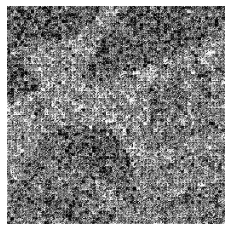}};
\node at (2*\xxspace,-13*\yspace) {\includegraphics[width=\iwidth]{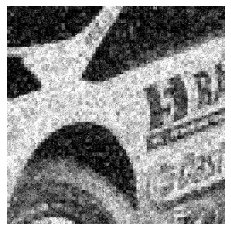}};
\node at (3*\xxspace,-13*\yspace) {\includegraphics[width=\iwidth]{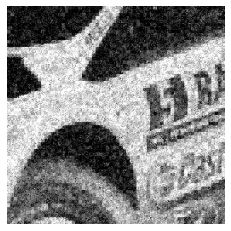}};
\node at (4*\xxspace,-13*\yspace) {\includegraphics[width=\iwidth]{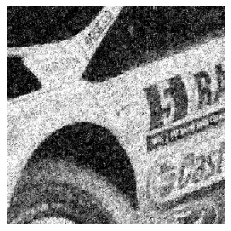}};
\node at (5*\xxspace,-13*\yspace) {\includegraphics[width=\iwidth]{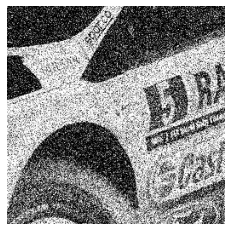}};
\node at (6*\xxspace,-13*\yspace) {\includegraphics[width=\iwidth]{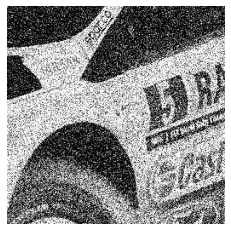}};
\node at (7*\xxspace,-13*\yspace) {\includegraphics[width=\iwidth]{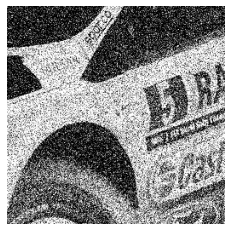}};

\node at (0*\xxspace,-11.2*\yspace) {\parbox{4cm}{\centering Iter.: 1\\ 5.2 dB}};
\node at (1*\xxspace,-11.2*\yspace) {\parbox{4cm}{\centering Iter.: 10\\  8.6 dB}};
\node at (2*\xxspace,-11.2*\yspace) {\parbox{4cm}{\centering Iter.: 50\\  16.0 dB}};
\node at (3*\xxspace,-11.2*\yspace) {\parbox{4cm}{\centering \textcolor{red}{Best Iter.: 61\\ 16.3 dB}}};
\node at (4*\xxspace,-11.2*\yspace) {\parbox{4cm}{\centering Iter.: 100\\  15.5 dB}};
\node at (5*\xxspace,-11.2*\yspace) {\parbox{4cm}{\centering Iter.: 700\\  12.5  dB}};
\node at (6*\xxspace,-11.2*\yspace) {\parbox{4cm}{\centering Iter.: 1100\\ 12.5 dB}};
\node at (7*\xxspace,-11.2*\yspace) {\parbox{4cm}{\centering Iter.: 1500\\ 12.5 dB}};

\node at (8.2*\xxspace,-13*\yspace) {\includegraphics[width=\iwidth]{./untrained/noisy_img.png}};
\node at (9.2*\xxspace,-13*\yspace) {\includegraphics[width=\iwidth]{./untrained/bm3d.png}};
\node at (10.2*\xxspace,-13*\yspace) {\includegraphics[width=\iwidth]{./untrained/clean.png}};

\node at (8.2*\xxspace,-11.2*\yspace) {\parbox{4cm}{\centering Noisy Image\\  12.5 dB}};
\node at (9.2*\xxspace,-11.2*\yspace) {\parbox{4cm}{\centering BM3D\\  23.5 dB}};
\node at (10.2*\xxspace,-11.2*\yspace) {\parbox{4cm}{\centering Clean Image}};

\end{tikzpicture}
}
\end{center}
\vspace{-0.3cm}
\caption{
\label{fig:examplesuntrained}
Visualizing the images obtained by fitting a single noisy image with different architectures. The U-Net fits the low-frequency components of the image before the high-frequency ones, while for the image-to-image mixer, we can see the structure imposed by mixing horizontally and vertically.
}
\end{figure}

\begin{table}[h]
\centering
\small
\begin{tabular}{l c c c c c }
\toprule
 & ImageNet & BSD68 & Urban100 & Kodak24 & McMaster18\\ 
\midrule 
Img2Img-Mixer& $\bm{0.9305}$& $\bm{0.9260}$& $\bm{0.9537}$& $\bm{0.9330}$& $\bm{0.9421}$\\ 
U-Net &0.9249 &0.9221 &0.9434 &0.9279 &0.9330\\ 
ViT &0.9194 &0.9178 &0.9352 &0.9221 &0.9229\\ 
Original-Mixer& 0.9136 &0.9106 &0.9289 &0.9143 &0.9209\\ 
BM3D & 0.8322 &0.8519 &0.8924 &0.8678 &0.8489 \\
\bottomrule
\end{tabular}
\caption{\label{tab:main_results_ssim} SSIM of the colour gaussian denoising task from Section \ref{sec:gaussiandenoising}. All networks have 24M parameters and are trained on 100k ImageNet images. Not only does the image-to-image mixer surpass all baselines with respect to the PSNR as shown in Table \ref{tab:main_results}, but also with respect to the SSIM.}
\end{table}

\begin{table}[h]
\centering
\begin{tabular}{l c c c c}
\toprule
Size & P & N & C & f \\
\midrule 
1.66M & 4 & 16 & 64 & 4 \\ 
2.40M  & 4 & 16 & 96 & 4 \\ 
3.44M  & 4 & 16 & 128 & 4 \\ 
6.61M  & 4 & 16 & 128 & 8 \\ 
12.19M & 4  &16 &  192 & 8 \\
24.18M & 4 & 16 & 400 & 4 \\
\bottomrule
\end{tabular}
\caption{\label{tab:param_config} Hyperparameter configuration for varying the network size of the image-to-image mixer in Figure ~\ref{fig:example}.}
\end{table}

\section{Further ablation studies}

\paragraph{Linear MLP-mixer layers.}
A perhaps very interesting variation of the image-to-image mixer works with linear mixer layers.  
Recall that the height mixing, width mixing, and channel mixing blocks all consist of a linear transformation, non-linearity, followed by another linear transformation. We studied a mixer version where instead of three such mixer layers we only have one mixer layer, which has a first linear layer mixing in height dimension, a second linear layer mixing in width dimension, a third linear layer mixing in channel dimension, followed by a non-linearity, and finally a linear layer in channel dimension. This mixer architecture is very simple and, perhaps surprisingly, performs almost as well as the default image-to-image mixer architecture introduced earlier. 
Specifically, we designed a 3M version with linear MLP-mixer layers and trained it on 4000 ImageNet images. It achieved 29.92 dB, only 0.15 dB less than the default Img2Img-Mixer of similar model size. 

\paragraph{Effect of patch size.}
Our default image-to-image mixer has a patch size of $P=4$ as in the Swin transformer~\citep{liu_SwinTransformerHierarchical_2021} and the Swin U-Net transformer~\citep{cao_SwinUnetUnetlikePure_2021}. 
Here, we evaluate versions of the network with varying patch sizes on the Gaussian denoising experiment described in Section~\ref{sec:gaussiandenoising} (4000 training images). We varied the patch sizes and changed the other hyperparameters to keep the model size similar. Table \ref{tab:patch_size} depicts the PSNR values for two model sizes. The results show that having a smaller patch size than 4 actually leads to a marginally better performance. However, a smaller patch size also leads to a higher computational cost, which is why $P=4$ is a good tradeoff.
The exact hyperparameter configurations are also in Table \ref{tab:patch_size}.

\begin{table}[h]
\centering
\small
\begin{tabular}{l c c }
\toprule
$P$  & 3.4M & 6.8M \\ 
\midrule 
1 & 30.19 & 30.26  \\ 
2 & 30.18 & 30.26  \\ 
4 & 30.07 & 30.20  \\ 
8 & 23.85 & 24.01  \\ 
\bottomrule
\end{tabular}
\quad
\hspace*{2cm}
\begin{tabular}{l c c c c}
\toprule
Size & P & N & C & f \\
\midrule 
3.45M  &1  &12 &107 & 1  \\ 
3.46M  & 2 & 16 & 140 &2 \\ 
3.44M  & 4 & 16 & 128 & 4 \\ 
3.46M & 8  &16 &  128 & 4 \\
\bottomrule
\end{tabular}
\quad
\begin{tabular}{l c c c c}
\toprule
Size & P & N & C & f \\
\midrule 
6.81M  &1  &12 &100 & 2  \\ 
6.82M  & 2 & 16 & 140 &4 \\ 
6.61M  & 4 & 16 & 128 & 8 \\ 
6.86M & 8  &16 &  140 & 8 \\
\bottomrule
\end{tabular}

\caption{\label{tab:patch_size} \textbf{Left:} Denoising PSNR (dB) for varying patch sizes $P$. Smaller patch sizes than 4 give marginally better numbers, but at the expense of higher computational cost. \textbf{Right:} Hyperparameters of the two model sizes. }
\end{table}

\section{Vision transformer for image reconstruction}
The original vision transformer~\citep{dosovitskiy2021an} was proposed to perform image classification tasks, and works as follows: It first partitions the input image into smaller image patches which are then linearly projected to a higher dimensional feature space. These patch embeddings, together with position embeddings and a classification token, are sent into a standard transformer encoder. At the the output of the transformer encoder, only the classification token is mapped to a class label via a classification head, which can be realized by an MLP.

In this work, we adapt the vision transformer to perform image reconstruction tasks by implementing two simple modifications: First, we discard the classification token as it becomes redundant for image reconstruction. Second, we replace the classification head by a reconstruction head that maps the transformer output back to a visual image.

The reconstruction head contains one layer normalization followed by a linear layer, which are shared across all the sequence elements of the transformer output. Hence, each sequence element in the feature space is mapped to a corresponding image patch in pixel space. The reconstructed image patches are then combined to a full-sized image.

\end{document}